\documentclass{article}
\usepackage[utf8]{inputenc}
\usepackage[T1]{fontenc}
\usepackage{tabularx}
\usepackage{amsfonts}
\usepackage{bm}
\usepackage{dsfont}
\usepackage{adjustbox}
\usepackage{algorithm}
\usepackage{algpseudocode}
\usepackage{graphicx}
\usepackage{hyperref}
\usepackage{amssymb,amsmath,array}
\newcommand{\indep}{\rotatebox[origin=c]{90}{$\models$}}

\title{Information-theoretic Bayesian Optimization: Survey and Tutorial}

\author{Eduardo C. Garrido-Merchán\\
Universidad Pontificia Comillas\\
  \texttt{ecgarrido@comillas.edu}}
\date{January 2025}

\begin{document}

\maketitle

\begin{abstract}
Several scenarios require the optimization of non-convex black-box functions, that are noisy expensive to evaluate functions with unknown analytical expression, whose gradients are hence not accessible. For example, the hyper-parameter tuning problem of machine learning models. Bayesian optimization is a class of methods with state-of-the-art performance delivering a solution to this problem in real scenarios. It uses an iterative process that employs a probabilistic surrogate model, typically a Gaussian process, of the objective function to be optimized computing a posterior predictive distribution of the black-box function. Based on the information given by this posterior predictive distribution, Bayesian optimization includes the computation of an acquisition function that represents, for every input space point, the utility of evaluating that point in the next iteraiton if the objective of the process is to retrieve a global extremum. This paper is a survey of the information theoretical acquisition functions, whose performance typically outperforms the rest of acquisition functions. The main concepts of the field of information theory are also described in detail to make the reader aware of why information theory acquisition functions deliver great results in Bayesian optimization and how can we approximate them when they are intractable. We also cover how information theory acquisition functions can be adapted to complex optimization scenarios such as the multi-objective, constrained, non-myopic, multi-fidelity, parallel and asynchronous settings and provide further lines of research.
\end{abstract}

\section{Introduction}
The fields of machine learning and artificial intelligence in general have gone through significant advancements in recent years, driven by the development of sophisticated algorithms, the rise of computing and the availability of extensive datasets. However, the successes of these algorithms often depend on the meticulous hyperparameter tuning process, that can be both time-consuming and computationally expensive, both in money and resources, being a cause of massive pollution \cite{brevini2020black}. Hyperparameter optimization, therefore, has emerged as a critical area of research to enhance the efficiency and performance of machine learning models.

Bayesian optimization is position as a state-of-the-art class of methods for hyperparameter tuning, particularly due to its ability to handle expensive black-box functions. A black-box function is expensive to compute, lacks an analytical expression and it is usually noisy. By leveraging probabilistic models like Gaussian processes or Bayesian neural networks, Bayesian optimization balances exploration and exploitation, efficiently identifying optimal hyperparameters with fewer evaluations than traditional methods, reducing the mentioned pollution. Originally proposed as a general-purpose optimization framework, it has found widespread application in machine learning tasks \cite{snoek2012practical}, but it can be also used successfully as a tool for automatic sequential design of experiments \cite{agrell2021sequential}.

Acquisition functions, central components of Bayesian optimization, determine the next point to evaluate by balancing exploration and exploitation based on the information of the probabilistic surrogate model. To ensure robust performance, it is desirable that these functions should (i) not only be based on heuristics but be rigorous, (ii) be global to all the information of the probabilistic surrogate model and not only in an evaluated point and (iii) be grounded in a formal mathematical theory with proven guarantees, like information theory \cite{cover1999elements}. Critically, information-theoretic approaches to Bayesian optimization provide a deeper theoretical understanding, as acquisition functions based on this approach satisfy the three previous statements. In particular, these methods use concepts such as entropy and mutual information to guide the optimization process more effectively, offering significant advantages over conventional approaches.

This paper aims to provide a comprehensive survey of information-theoretic Bayesian optimization, bridging the gap between theoretical advancements and practical implementations. In particular, information theoretical acquisitions can be very complicated technically and in recent years lots of approaches based on these concepts have been published. Consequently, we believe that this paper is very significant as it serves as a (i) historical review of information theory based Bayesian optimization (ii) tutorial of the approaches to read the technical papers of every one of them more easily and (iii) survey of all the approaches that have been published recently. Concretely, we explore the foundational principles, key methodologies, and state-of-the-art techniques in this domain. By doing so, we aim to guide researchers and practitioners with the knowledge to leverage these methods in their respective applications successfully.

We begin this paper with an overview of Bayesian optimization fundamentals that serves as a tutorial for the rest of the content. Then, we continue with a brief section that summarizes the main information theory applied concepts in Bayesian optimization, which are necessary to understand the following section, where we describe in detail the main information theoretical based approaches to Bayesian optimization and how are they applied to advanced scenarios such as taking into account multiple objectives and constraints. Finally, the last section describes the main conclusions and further lines of research of information theoretical based Bayesian optimization.  
\section{Fundamentals of Bayesian optimization}
As we have stated in the previous section. Bayesian optimization is a probabilistic model-based approach to optimize expensive black-box functions. Its theoretical foundation, initially proposed by Mockus \cite{movckus1975bayesian}, provides a structured framework for sequentially selecting configurations of a black-box problem, such as hyper-parameter values in the hyper-parameter tuning problem, to minimize (or equivalently maximize) an unknown objective black-box function,$f(\mathbf{x})$. More formally, we seek to obtain an approximation to $\mathbf{x}^\star$ evaluating as less configurations $\mathbf{x}$ as possible, as every different evaluation of $f(\mathbf{x})$ is costly:
\begin{align}
\mathbf{x}^\star = argmin_{\mathbf{x} \in \mathcal{X}}(f(\mathbf{x}))\,,
\end{align}
where $\mathcal{X}$ is the configuration input space. In order to solve this global non-convex optimization problem, Bayesian optimization uses two key components: a probabilistic surrogate model $M(\mathcal{D})$ that is fitted according to the dataset of observed configurations $\mathcal{D} = \{\mathbf{X}, \mathbf{y}\}$ and an acquisition function $\alpha(M(\mathbf{x}))$, which is a criterion that uses the information given by the probabilistic model $M(\mathbf{x})$ to estimate how useful is to evaluate a point of the input space $\mathbf{x} \in \mathcal{X}$ in order to gain information about the optimum of the problem $\mathbf{x}^\star$ in the following iteration using a trade-off between exploration and exploitation. Both components must be chosen by the practitioner, being the most popular choices for the probabilistic surrogate a Gaussian process model \cite{williams2006gaussian}, a random forest or a Bayesian neural network. In the case of acquisition functions, the most popular choices are the expected improvement, the probability of improvement or information theoretical acquisition functions, which are the topic of this manuscript.

As the most popular model choice in Bayesian optimization, Gaussian processes offer a flexible representation to the unknown objective function $f(\mathbf{x})$, being non-parametric models that are equivalent, under certain assumptions, to neural networks and configurable through kernel functions $k(\mathbf{x}, \mathbf{x}')$, defining a functional space $\mathcal{F}$ where we assume that the objective function belongs to $f(\mathbf{x}) \in \mathcal{F}$. In other words, they are non-parametric probabilistic models used to infer a distribution over functions $\mathcal{F}$, or more formally, a collection of random variables, any finite subset of which follows a joint Gaussian distribution. More formally, a Gaussian process is defined by its mean function \( m(\mathbf{x}) \) and covariance function \( k(\mathbf{x}, \mathbf{x}') \), written as:
\begin{align}
f(\mathbf{x}) \sim \mathcal{GP}(m(\mathbf{x}), k(\mathbf{x}, \mathbf{x}'))\,,
\end{align}
where \( m(\mathbf{x}) = \mathbb{E}[f(\mathbf{x})] \) is the expected value of the function and \( k(\mathbf{x}, \mathbf{x}') = \text{Cov}(f(\mathbf{x}), f(\mathbf{x}')) \) describes the covariance between the function values at two points, controlling smoothness and variability.

If the assumption about the objective function $f(\mathbf{x})$ belonging to the distribution of functions $\mathcal{F}$ is satisfied, $f \in \mathcal{F}$, then, using a Gaussian process will guarantee that the Bayesian optimization procedure is successful. Prior knowledge about the objective function can be represented through the prior mean $m(\mathbf{x})$ of the Gaussian process, the choice of covariance function $k(\mathbf{x}, \mathbf{x}')$ and its hyperparameters $\theta \in \Theta$, like the lengthscales of the covariance function $\mathbf{l}$.

The predictive distribution used by the Gaussian process model is computed in the following way given the observations of the configurations already evaluated. Given observed configurations \( \mathcal{D} = \{(\mathbf{x}_i, y_i)\}_{i=1}^n \), where \( y_i = f(\mathbf{x}_i) + \epsilon \) (with noise \( \epsilon \sim \mathcal{N}(0, \sigma_n^2) \)), a Gaussian process model produces a posterior predictive distribution, that is used to compute the acquisition function. for any test point \( \mathbf{x}_* \):
\begin{align}
f(\mathbf{x}_*) \mid \mathcal{D} \sim \mathcal{N}(\mu(\mathbf{x}_*), \sigma^2(\mathbf{x}_*))\,,
\end{align}
where the predictive mean \( \mu(\mathbf{x}_*) \) and variance \( \sigma^2(\mathbf{x}_*) \) are given by:
\begin{align}
\mu(\mathbf{x}_*) = \mathbf{k}_*^\top [\mathbf{K} + \sigma_n^2 \mathbf{I}]^{-1} \mathbf{y},
\end{align}
\begin{align}
\sigma^2(\mathbf{x}_*) = k(\mathbf{x}_*, \mathbf{x}_*) - \mathbf{k}_*^\top [\mathbf{K} + \sigma_n^2 \mathbf{I}]^{-1} \mathbf{k}_*.
\end{align}
such that \( \mathbf{K} \) is the \( n \times n \) covariance matrix of the observed inputs, with entries \( K_{ij} = k(\mathbf{x}_i, \mathbf{x}_j) \) and \( \mathbf{k}_* \) is the covariance vector between \( \mathbf{x}_* \) and the observed data points.

Given this predictive distribution, that represents our beliefs about the objective function value $f(\mathbf{x})$ for every different point $\mathbf{x}$ of the input space $\mathcal{X}$, the Bayesian optimization algorithm computes an acquisition function $\alpha(\mathbf{x})$, that represents the expected utility of evaluating every different point $\mathbf{x} \in \mathcal{X}$ with the purpose of it being the optimum of the problem $\mathbf{x}^\star$. The most common choice is the expected improvement acquisition function, due to its simplicity, although it has been proven empirically that entropy based acquisition functions tend to outperform this criterion in a wide array of problems \cite{hernandez2014predictive}. The expected improvement is just a local criterion that represents the expectation $\mathbb{E}[\cdot]$ of the improvement $I(\mathbf{x})$ that an input point $\mathbf{x}$ is going to have over the best point seen so far, or incumbent, $\xi$. The improvement of a point $\mathbf{x}$ is defined as $I(\mathbf{x}) = max(f(\mathbf{x} - \xi, 0))$, where we compute $f(x)$ in practice as the prediction done by the Gaussian process $\mu(\mathbf{x})$. We can compute the probability of improvement, hence, including the uncertainty about the mean of the Gaussian process such that $f(\mathbf{x}) \approx \mathcal{N}(\mu(\mathbf{x}), \sigma^2(\mathbf{x}))$. Then, to estimate the probability of improvement we can just compute the right tail of the cumulated function of the normal distribution $\Phi(\mathbf{x})$ in the point $\mathbf{x}$ above from the best seen point so far $\xi$, which makes the probability of improvement acquisition function given by the following expression:
\begin{align}
PI(\mathbf{x}) = 1 - \Phi(\frac{f(\mathbf{x}^\star) - \mu(\mathbf{x})}{\sigma(\mathbf{x})}) \,.
\end{align}
Then, the expected improvement is just the expectation of the previous random variable taken with respect to the posterior predictive distribution provided by the Gaussian process, which after some algebra has the following closed form analytical expression:  
\begin{align}
EI(\mathbf{x}) = (\mu(\mathbf{x}) - f(\mathbf{x}^\star))\Phi(\frac{\mu(\mathbf{x}) - f(\mathbf{x})}{\sigma(\mathbf{x})}) + \sigma(\mathbf{x})\phi(\frac{\mu(\mathbf{x}) - f(\mathbf{x})}{\sigma(\mathbf{x}})\,, 
\end{align}
that can be enhanced to enforce exploration, as the criterion tends to exploit, as it is based on the local information that gives the predictive distribution of every single different point $\mathbf{x}$, not consider all the predictive distribution. Hence, despite the fact that the expected improvement criterion is computationally efficient and conceptually simple, it has several notable limitations: It tends to exploit more than explore, as it is based on local information, not taking into account regions where the uncertainty is high, hence, as dimensionality increases, the balance between exploration and exploitation becomes harder to achieve, reducing its performance. Consequently, if a neighbourhood of an optimum has not been explored, expected improvement may fail to adequately explore it, becoming worse when multiple promising regions exist. 

\begin{figure}[htb!]
  \centering
  \includegraphics[width=0.99\textwidth]{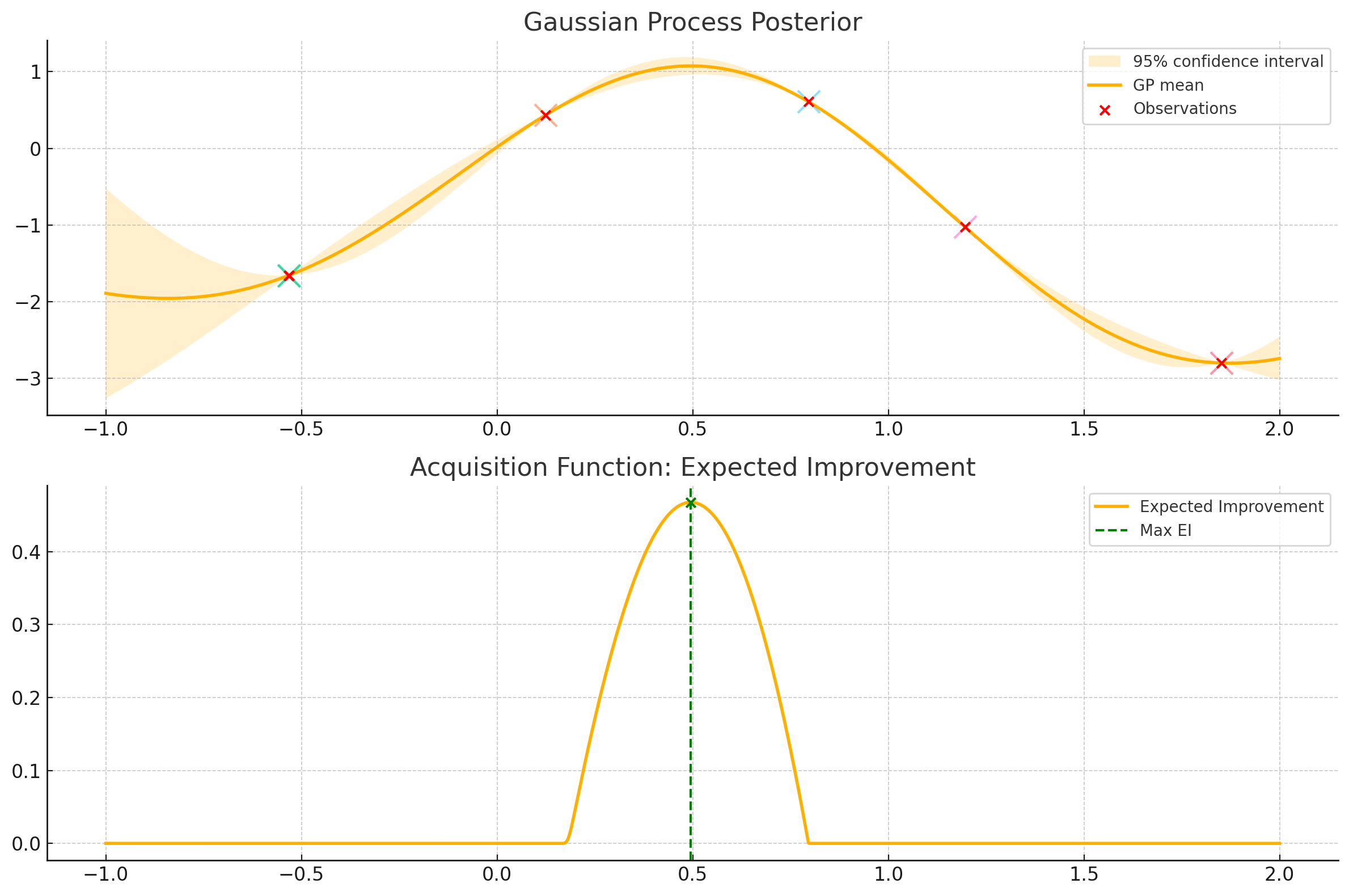}
	\caption{Gaussian process posterior distribution of an objective function and the associated acquisition function (Expected Improvement \cite{garnett2023bayesian}) whose maximum represents the following suggestion of the Bayesian optimization procedure. We can see how a better prediction incur in a higher value of the acquisition, that is even higher if the associated uncertainty is higher. However uncertainty on the left is not taken into account by Expected improvement, and the optimum may be there. }
  \label{fig:gp_af}
\end{figure}

On the other hand, information theoretical based criteria take into account the uncertainty of all the predictive distribution of the model, with the purpose of maximizing the information gain in every iteration of some random variable related with the optimum of the problem. Consequently, this approach does not present the issues of local criteria similar to the expected improvement criterion. This is the reason why we believe that it is critical to explain information theoretical based criteria in this survey. We visualize the explained probabilistic surrogate model, in this case a Gaussian process, and the expected improvement acquisition function in Figure \ref{fig:gp_af}.
The Bayesian optimization vanilla algorithm uses the previous components sequentially to decompose the difficult global non-convex optimization problems into sequential simple optimization problems where the maximum of the acquisition function $\alpha(\mathbf{x})$ is the suggested point for evaluation in the black-box $f(\mathbf{x})$. Importantly, optimizing the acquisition function is cheap as gradients can generally be computed and the evaluation is cheap as it only needs the predictive distribution of the Gaussian process. Once a new observation is evaluated, it is used to condition the Gaussian process, acquiring additional information of the objective function that is useful to guide the search towards the optimum. Once a budget of iterations is finished, Bayesian optimization recommends as the solution of the problem the best point observed so far or the point that optimizes the Gaussian process. We summarize the previous steps in Algorithm \ref{alg:bayesian_optimization}. 
\begin{algorithm}
\caption{Bayesian Optimization Basic Algorithm}
\label{alg:bayesian_optimization}
\begin{algorithmic}[1]
	\State \textbf{Input:} Observation Dataset $\mathcal{D}_0 = \{(\mathbf{x}_i, y_i)\}_{i=1}^{n_0}$, GP model $p(f)$, 
	acquisition function $a(\cdot)$, number of iterations $T$
\State Fit Gaussian process model to $\mathcal{D}_0$
\For {$t = 1$ to $T$}
	\State Select recommendation point $\mathbf{x}_t = \arg\max_{\mathbf{x}\in\mathcal{X}} a(\mathbf{x})$
	\State Evaluate the objective function $y_t = f(\mathbf{x}_t) + \epsilon_t$, where $\epsilon_t \sim \mathcal{N}(0, \sigma^2)$
	\State Add it to the dataset $\mathcal{D}_t = \mathcal{D}_{t-1} \cup \{(\mathbf{x}_t, y_t)\}$
    \State Condition Gaussian process model with $\mathcal{D}_t$
\EndFor
	\State \textbf{Output:} The best point found $\mathbf{x}_i$, where $i = \arg\max  y_i$.
\end{algorithmic}
\end{algorithm}

Explaining Bayesian optimization in every single small technical detail is out of the scope of this paper. For a complete explanation of the Bayesian optimization class of methods we recommend the monographic book by Garnett, which covers a wide array of concepts regarding Bayesian optimization \cite{garnett2023bayesian}.

\section{Information Theory Applied Concepts in Bayesian optimization}
Having seen the fundamentals of Bayesian optimization we continue this work with a short but rigurous exposition of basic information theory concepts \cite{cover1999elements} that must be understood to comprehend the information theory Bayesian optimization approaches that will be explained in detail in the following section. 

Information theory was pioneered by Claude E. Shannon in his seminal paper "A Mathematical Theory of Communication" \cite{shannon1948mathematical}. In this work, Shannon introduced foundational concepts that revolutionized communication and data transmission. Based on that exposition, we extract the concepts that are mostly used for Bayesian optimization. 

\begin{figure}[htb!]
  \centering
  \includegraphics[width=0.99\textwidth]{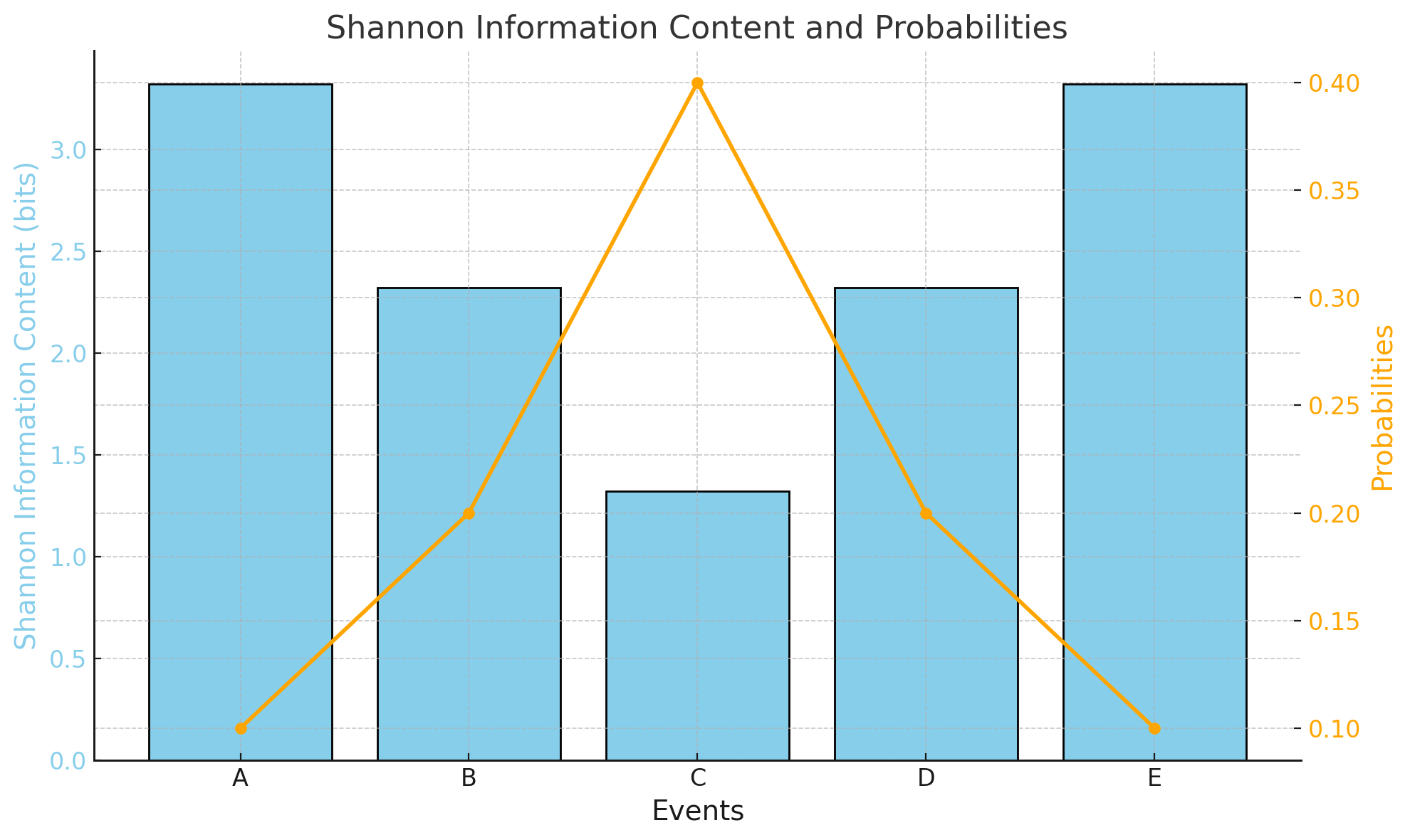}
	\caption{Shannon information content (blue) and probabilities (yellow) associated with the 5 events of a discrete random variable. If probabilities are lower, the surprise of the event is higher, represented by the Shannon information content and viceversa.}
  \label{fig:sic}
\end{figure}

\begin{figure}[htb!]
  \centering
  \includegraphics[width=0.99\textwidth]{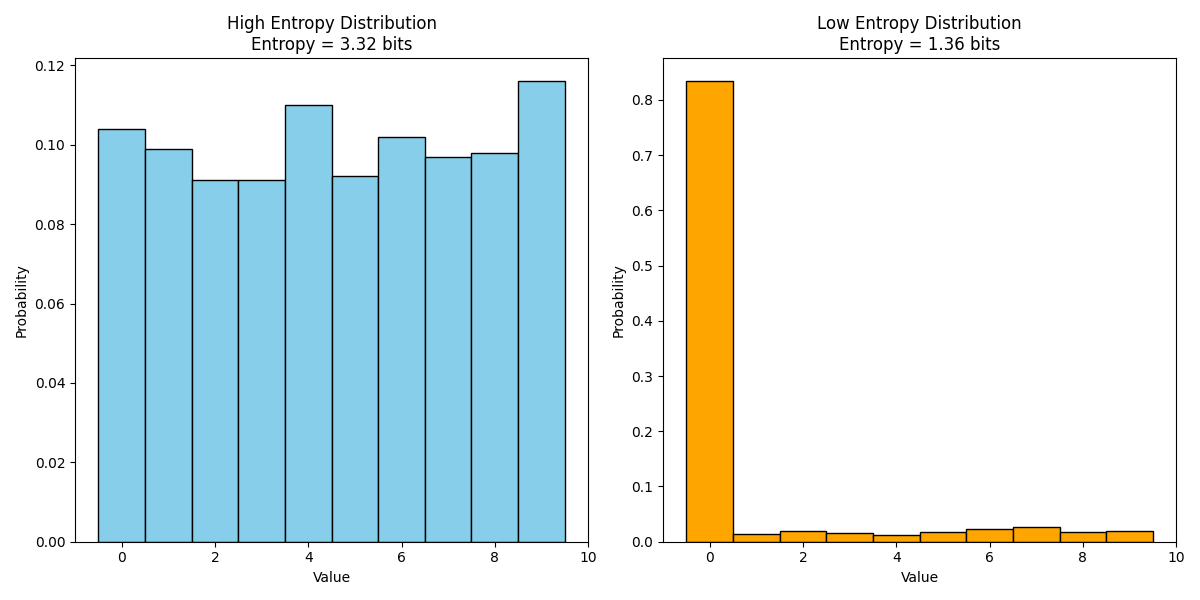}
	\caption{Entropy of two discrete random variables. The entropy is the expectation of the surprise of the events of a random variable whose probability is given by the mass probability function. The random variable of the left has higher entropy than the one on the right. This means that we have more certainty about the values of the random variable of the right, as the lowest one is more likely to happen whereas on the left we do not know which will be the next outcome. Consequently, we require less bits to encode its values. This will be used for information theoretical Bayesian optimization approaches as we will discuss further.}
  \label{fig:entropy}
\end{figure}

We begin with the Shannon information content of an event $x$ of a random variable $X$, with probability $p(X=x)$, being defined as:
\begin{align}
SIC(x) = \log_2 \frac{1}{p(x)},
\end{align}
where $SIC(x)$ is measured in bits if the logarithm is base 2. This quantity represents the surprise of the outcome $x$ as a result of an experiment of the random variable X. We illustrate in Figure \ref{fig:sic} how events with higher probability have a lower surprise.  
If surprises of the events were lower in average, then, the outcomes of the variable $X$ are going to be more predictable and viceversa, so we can interpret that we have more knowledge about one random variable if the expectation of the Shannon information content of its events is lower than the one of another variable that is higher. This expectation of the Shannon information content is known as the entropy of a discrete random variable $X$, $\mathbb{E}(SIC(X)) = H(X)$, as we visualize in Figure \ref{fig:entropy}, where we see how the entropy can be used to measure our knowledge about the values that will result as an outcome of a random variable.
More formally, the entropy is given by the following expression:
\begin{align}
H(X) = \sum_{x \in X} p(x) \log \frac{1}{p(x)} = - \sum_{x \in X} p(x) \log p(x),
\end{align}
where the sum is substituted with an integral in the case of continuous random variables, expression that is known as the differential entropy of a continuous random variable $X$, and that uses its probability density function:
\begin{align}
H(X) = -\int p(x) \log p(x) dx,
\end{align}
As the entropy is computed wrt a random variable, it could be from a multivariate random variable or from a joint distribution, for example let X and Y be random variables then the entropy of the joint random variable is $H(X,Y) = - \int p(x, y) \log p(x, y) dx dy$.  

The previous result is particularly useful for Bayesian optimization, as it is usually interesting to know whether an observation gives you information about another random variable, such as the optimum of a problem. Concretely, if two variables $X$ and $Y$ are independent, then its joint distribution factorizes, $p(X,Y)=p(X)p(Y)$, entropy is additive and no information about $X$ is obtained if $Y$ is known. However, if $Y$ gives information about $X$, then the variables are dependent and $p(X,Y)=p(X|Y)P(Y)$ holds. To measure how much information does $Y$ give about $X$, we can use the mutual information expression $I(X,Y)$, which can be interpreted as the distance between the joint probability distribution of the variables and the information given by their marginal distributions. Consequently, it measures the information of the dependency between $X$ and $Y$, or how much knowing about one variable reduces the uncertainty about the other, and it is given by the following expression: 
\begin{align}
I(X,Y) = \int \int p(X,Y) \log \frac{p(X,Y)}{p(X)p(Y)}dx dy\,,
\end{align}
where $I(X,Y)=0$ if $X \indep Y$, being non-negative and symmetric. The amount of knowledge that we obtain about one random variable knowing the other is also referred as information gain, which is going to be useful afterwards. Interestingly, this quantity is also related with the entropy of a conditional distribution $H(X|Y)$, what is known as the conditional differential entropy $H(X|Y) = \int \int p(x, y) log p(x|y) dx dy$. More concretely, it can be derived that the mutual information is equal to the entropy of the marginal distribution of $X$ minus the entropy of the conditional distribution $I(X,Y)=H(X) - H(X|Y) = H(Y) - H(Y|X)$, what can be intuitively interpreted as the gain of knowledge that we obtain about $X$ when we condition the random variable $X$ to another variable $Y$, if both are independent $X \indep Y$ we obtain that $I(X,Y)=H(X) - H(X) = 0$. 

Finally, the previous expression can also be interpreted as the Kullback Leibler divergence $D_{KL}(p(X,Y)||p(X)p(Y))$ between the joint distribution $p(X,Y)$ of two random variables and its factorization into marginal distributions $p(X)p(Y)$. The explained concepts represent only a small and necessary fraction of the information theory field that are mandatory to understand the approaches that are going to be illustrated in the following section. For a complete overview of information theory, covering a wider array of concepts that the ones explained in this section and specifically applied to machine learning related tasks, we recommend the book by David J.C. Mackay \cite{mackay2003information}.
\section{Information theoretical based approaches to Bayesian Optimization}
Having covered the fundamentals of Bayesian optimization and information theory we now describe in detail the information-theoretic based approaches to Bayesian optimization in chronological order, to justify their motivation.  

The first work on informational theoretic Bayesian optimization that we will include in the chronology of this topic was the informational approach to global optimization (IAGO) \cite{villemonteix2009informational} algorithm. This method combined the existing stepwise uncertainty reduction (SUR) algorithm \cite{geman1996active} with the previous efficient global optimization (EGO) algorithm \cite{jones1998efficient}, that used the expected improvement acquisition function with probabilistic surrogate models. As SUR selects the evaluation that maximizes the information gain that a potential new evaluation obtains about the optimum of the problem, acting as activation function $\alpha(\mathbf{x})$, and EGO selects the observation whose activation function is maximized and conditions the Gaussian process on the observed value $y=f(\mathbf{x}) + \epsilon$ such that $\epsilon = \mathcal{N}(0, \sigma)$ and $\mathbf{x} = argmax_{\mathbf{x} \in \mathcal{X}} \alpha(\mathbf{x})$, then, we can argue with certainty that this is the first approach that combined information-theoretic concepts with Bayesian black-box optimization. 

\begin{figure}[htb!]
  \centering
  \includegraphics[width=0.99\textwidth]{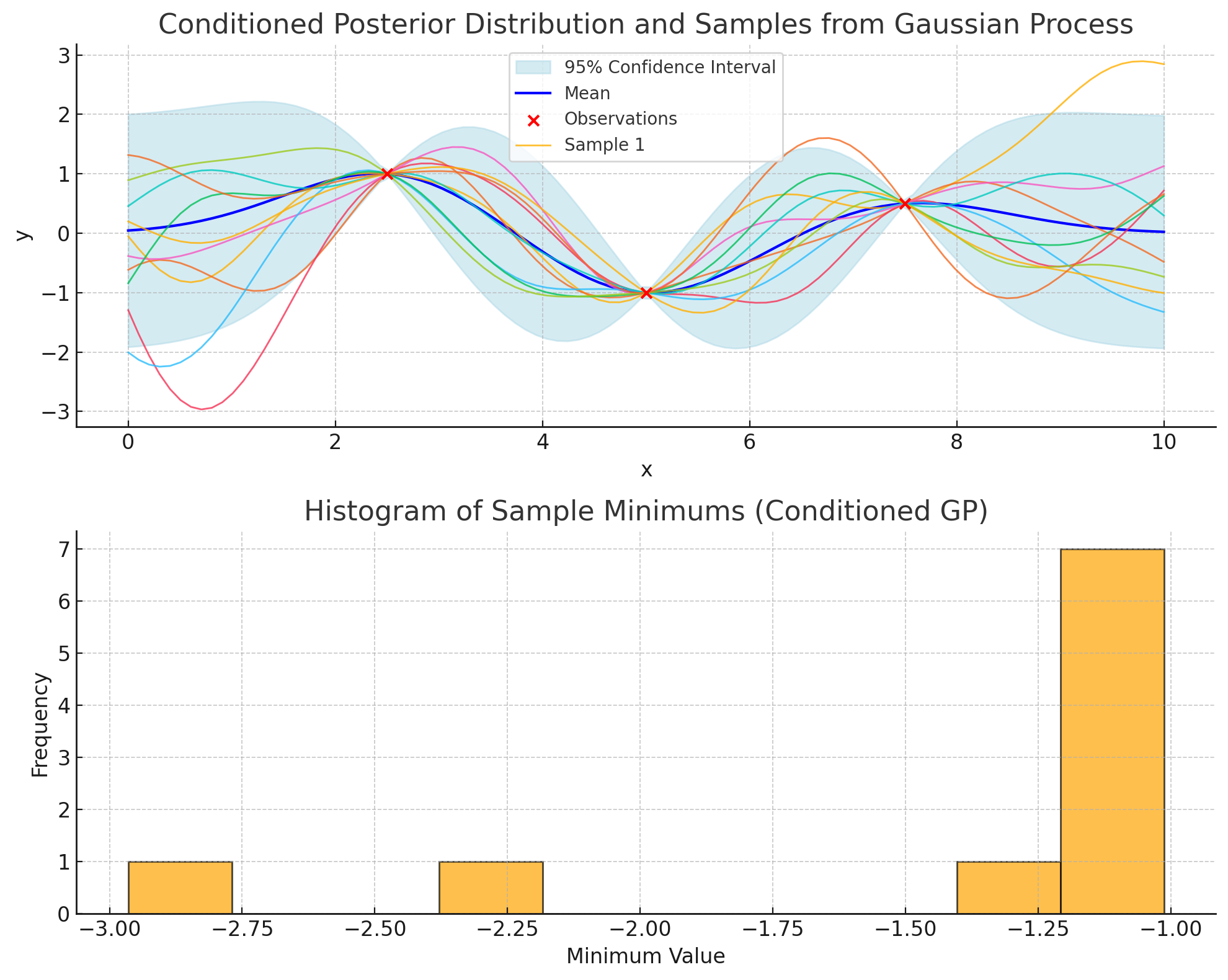}
	\caption{Empirical distribution of the minimizer $p(\mathbf{x}^\star)$ of a black-box function according to the information given by the posterior distribution of the conditioned Gaussian process on previous observations. Each sample consists on a path of the Gaussian process model that is optimized to obtain a sample of the minimum of the problem given the current information.}
  \label{fig:px}
\end{figure}

The main challenge of IAGO was estimating the density of $\mathbf{x}^\star$, the solution of the problem, in order to compute the information gain that any evaluation $\mathbf{x}$ would incur in, according to its predicted value by the posterior distribution of a Gaussian process model conditioned on a set of initial observations of the objective function $\mathcal{D} = \{(\mathbf{X}, \mathbf{y})\}$. To do so, it uses several samples from the conditioned Gaussian process model, optimizes them and builds an empirical distribution of a sum of delta functions, building the probability distribution of $\mathbf{x}^\star$.
\begin{align}
p(\mathbf{x}^\star|\mathcal{D}) = \frac{1}{s}\sum_{i=1}^s\delta_{x^\star_i}(\mathbf{x})\,,
\end{align}
where $s$ is the number of Monte Carlo samples. We illustrate in Figure \ref{fig:px} this process and the details of the conditioning, optimization and sampling can be found in the paper \cite{villemonteix2009informational}.
Being able to compute $p(\mathbf{x}^\star|\mathcal{D})$ imply that we can observe changes in $p(\mathbf{x}^\star|\mathcal{D})$ when we condition the Gaussian process with its prediction $\mathbf{y}$ of any input space point $\mathbf{x} \in \mathcal{X}$, obtaining as a different random variable $p(\mathbf{x}^\star|\mathcal{D} \cup (\mathbf{x}, \mathbf{y}))$. Hence, we measure the expected utility of evaluating a new point with the expected information gain of the conditioned probability of the optimum in a new point $p(\mathbf{x}^\star|\mathcal{D} \cup (\mathbf{x}, \mathbf{y}))$ with respect to the unconditioned distribution $p(\mathbf{x}^\star)$. Following the concepts introduced in the previous section, we can estimate the expected information gain of any input space point:
\begin{align}
I(p(\mathbf{x}^\star|\mathcal{D}), p(\mathbf{x}^\star|\mathcal{D} \cup (\mathbf{x}, \mathbf{y}))) = H(p(\mathbf{x}^\star|\mathcal{D})) - H(p(\mathbf{x}^\star|\mathcal{D} \cup (\mathbf{x}, \mathbf{y})))\,,
\end{align}
being the maximum of the previous acquisition function the input space point with highest expected information gain, being the recommendation of IAGO for being evaluated. In order to compute the conditional entropy $H(p(\mathbf{x}^\star|\mathcal{D} \cup (\mathbf{x}, \mathbf{y})))$, IAGO executes an empirical mean of a discrete conditional entropy through simulations using a quantization operator $Q$, hence discretizing the image space $f_Q(X) = Q(f(X))$. It basically discretizes the space by a finite set of $M$ real numbers, avoiding the need to estimate the differential conditional entropy \cite{villemonteix2009informational}. However, this is a costly procedure for multi-dimensional problems, as the computational cost of keeping the same accuracy when dimensionality rises leads to a exponential increase of complexity, being desirable to directly minimize the entropy in the original input space $\mathcal{X}$.   

Soon after, the Entropy Search \cite{hennig2012entropy} acquisition function was suggested to circumvent some of the issues of the previous approach. This method also suggests the recommended input space point $\mathbf{x} \in \mathcal{X}$ as a result of maximizing the information gain about the optimizer in every iteration, as IAGO did. However, the method to obtain the information gain is different, as Entropy Search models the optimum $\mathbf{x}^\star$ of the problem using a functional definition:
\begin{align}
p(\mathbf{x}^\star) = p(\mathbf{x}^\star = argmin f(\mathbf{x})) = \int_{\mathcal{F}}p(f)\prod_{\mathbf{x} \neq \mathbf{x}^\star} \theta(f(\mathbf{x}^\star) - f(\mathbf{x})) df\,,
\end{align}
such that the $\theta$ factors are heaviside functions, hence being one only if the point is the extremum and 0 otherwise. As the previous expression is the expectation on the functional space $\mathcal{F}$ defined by the Gaussian process, we find that the previous expression is equivalent to the IAGO expression but considering that certain functions have more probability than others in the functional space, which is more rigurous theoretically than just formalizing the modelization of the extremum by the samples, although in practice are quite similar as $p(f)$ is set by the distribution of functions of the Gaussian process and the integral is approximated by Monte Carlo sampling.

The authors of Entorpy Search suggest also a discretization of the input space but placing weights on areas such that a high resolution on them, having more discretization points, has a strong influence on the change of value of the prediction, which is done through expected improvement. This alleviates the high-dimensionality problem, as non-important regions of the space are weakly sampled and high-variance regions are sampled more to inspect whether they lead to more information gain, but it is also another issue of entropy search, as it is also computationally expensive.

As the probability of the functions $p(f)$ is a Gaussian process and the space is discretized, the estimation of the optimum can be substituted with the following expression:
\begin{align}
p(\mathbf{x}^\star) = \int_{f \in \mathbb{R}^n}\mathcal{N}(f|\mu, \Sigma)\prod_{\mathbf{x} \neq \mathbf{x}^\star} \theta(f(\mathbf{x}^\star) - f(\mathbf{x})) df\,,
\end{align}
such that $N$ are the discrete locations obtained in the discretization and the prior over functions $p(f)$ is a multivariate normal given by the Gaussian process evaluated on those points. However, the previous integral is intractable, but the authors solve it through an approximation $q_{min}$ given by the variational inference expectation propagation algorithm. If you are not familiar to these complex approximation methods to integrals we refer to a variational inference tutorial \cite{blei2017variational} and a expectation propagation exposition \cite{minka2001family}. Although the approximation is effective, its computational complexity $\mathcal{O}(N^4)$ incurs in problems in high-dimensionality scenarios, as the resolution of the space will be low in these settings. Finally, the previous modelization is used to obtain the information gain of every possible point in the input space using a first-order approximation of the expected change in the entropy of $p(\mathbf{x}^\star)$, as one advantage of the expectation propagation approximation $q_{min}$ to $p(\mathbf{x}^\star)$ is that it provides analytic derivatives, that can be used to optimize the acquisition function. For more technicalities on Entropy Search, we refer the reader to its paper \cite{hennig2012entropy} and to the Gaussian probabilities and the Expectation Propagation algorithm paper that explains the steps followed in the approximation of the Gaussian integral and the heaviside factors \cite{cunningham2011gaussian}.

ES is a powerful yet computationally very expensive approach that struggles in high dimensionality. To alleviate this issue, the Predictive Entropy Search (PES) acquisition function was proposed \cite{hernandez2014predictive}, that also uses the information gain of a candidate $\mathbf{x}$ towards the optimum $\mathbf{x}^\star$ but that considers the fact that the information gain metric between a pair of random variables $X,Y$ is symmetric $I(X,Y)=I(Y,X)$, which is going to simplify the problem of approximating the information gain. Concretely, PES frames the problem as recommending $\mathbf{x}_{n+1}$ such that it is the point that maximizes the expected reduction of the negative differential entropy of the distribution of the optimum $p(\mathbf{x}^\star|\mathcal{D})$ with respect to the entropy of the expected value predicted by the Gaussian process at that point $p(y|\mathcal{D}, \mathbf{x})$:
\begin{align}
\alpha(\mathbf{x}) = H(p(\mathbf{x}^\star|\mathcal{D})) - \mathbb{E}_{p(y|\mathcal{D}, \mathbf{x})}(H(p(\mathbf{x}^\star|\mathcal{D} \cup \{\mathbf{x}, y\})))\,,
\end{align}
where we can use the symmetric property of the mutual information to obtain the following equivalent acquisition function, which is the one approximated by predictive entropy search:
\begin{align}
\alpha(\mathbf{x}) = H(p(y|\mathcal{D}, \mathbf{x})) - \mathbb{E}_{p(\mathbf{x}^\star|\mathcal{D})}(H(p(y|\mathcal{D}, \mathbf{x}, \mathbf{x}^\star)))\,,
\end{align}
where $p(y|\mathcal{D}, \mathbf{x})$ is the posterior predictive distribution of $y$ given by the conditioned Gaussian process and the location of the global maximizer. Now, approximations are easier than in the previous case as the first term $H(p(y|\mathcal{D}, \mathbf{x}))$ is just the entropy of a normal distribution plus the noise of the evaluation. The second term, the entropy of the conditional probability distribution $H(p(y|\mathcal{D}, \mathbf{x}, \mathbf{x}^\star))$, however, still demands to use the expectation propagation algorithm to be approximated, where the expression being approximated contains several factors that are used to condition the distribution to the fact that $\mathbf{x}^\star$ is the optimum of the problem. If the reader is interested in those details and how the factors are incorporated into the conditional probability distribution, they are available in the supplementary material of the predictive entropy search paper \cite{hernandez2014predictive}.   
As several acquisition functions  have been introduced and depending on the problem some outperform others, an entropy search portfolio \cite{shahriari2014entropy} was proposed as a meta policy $u(\mathcal{X|\mathcal{D}})$ or meta acquisition function, that consists on a portfolio of acquisition function that is motivated by information theoretic considerations. Concretely, it basically consists on determining which of the recommendations $\mathbf{x} \in \mathcal{X}$ made by a set of acquisition functions $\boldsymbol{\alpha}(\mathcal{X})$ minimizes the most the expected reduction of entropy of the minimizer $p(\mathbf{x}|\mathcal{D})$:
\begin{align}
u(\mathcal{X|\mathcal{D}}) = argmin_{\mathbf{x}_1, ..., \mathbf{x}_n} \mathbb{E}_{p(y|\mathcal{D}, \mathbf{x})} (H(p(\mathbf{x}^\star|\mathcal{D} \cup \{\mathbf{x}, y\})))\,,
\end{align}
where $n$ is the countable size of the set of acquisition functions $\boldsymbol{\alpha}(\mathcal{X})$ of the portfolio, so intuitively we are selecting in each iteration the acquisition function that minimizes the expected entropy about the minimizer. Interestingly, this meta-policy could be applied to any number of information-theoretic acquisitions and other acquisitions.

As information-theoretic approaches were being proposed, the first customizations of the method were also beggining to appear. Although they are explained in Section 4.2.2, in 2016, the previously described ES approach was adapted for robotics to make a robot in an unknown environment collect its next measurement at the location estimated to be most informative within its current field of view, computing mutual information across the state space of the robot \cite{bai2016information}, in an analogy with reinforcement learning but in a lower space. This customization shows the adaptive potential of information-theoretic Bayesian optimization to any type of scenarios.  

\begin{figure}[htb!]
  \centering
  \includegraphics[width=0.99\textwidth]{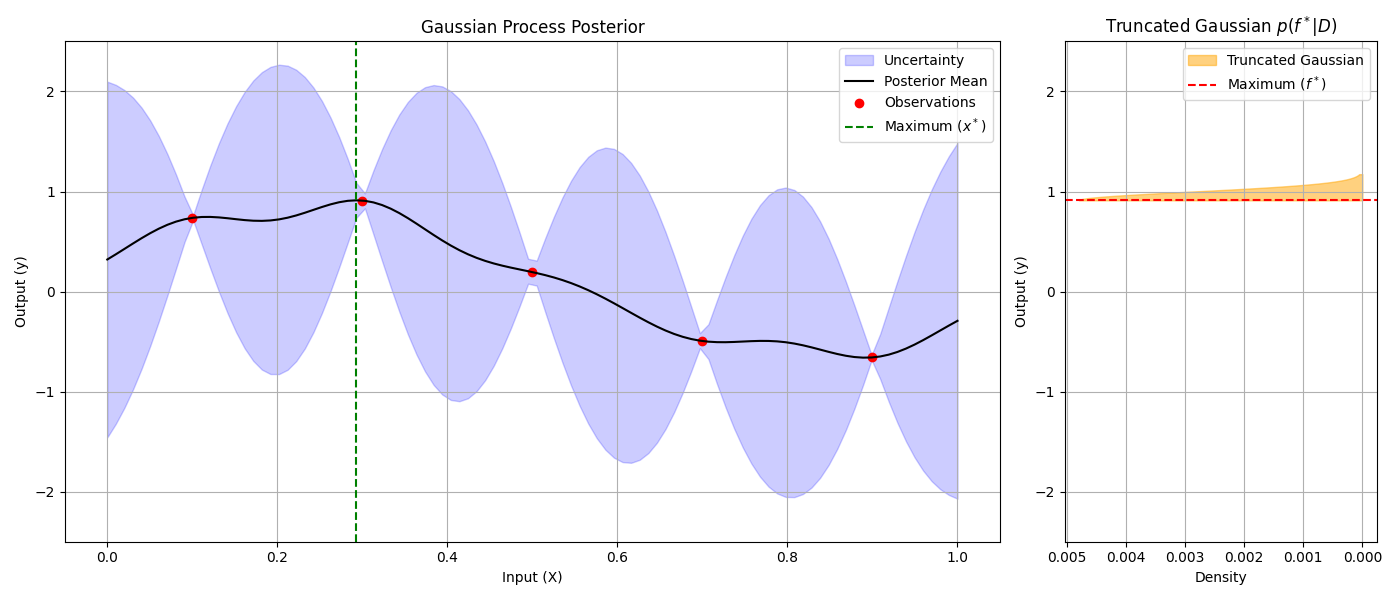}
	\caption{Visual example of the distribution of the optimum value of the problem given the information provided by the predictive distribution of the Gaussian process conditioned on observations $p(f^\star|\mathcal{D})$.}
  \label{fig:trun}
\end{figure}

Instead of considering the location of the optimum $p(\mathbf{x}^\star)$ as the random variable whose information is maximized through iterations, Max-value entropy search (MES) \cite{wang2017max} considers instead to maximize the expected information gain about the value of the black-box function at that location, $p(y)$, drastically simplifying the problem, as the black-box function satisfies $f : \mathbb{R}^n \to \mathbb{R}$, what makes the distribution of the optimum value $p(y)$ univariate. More formally, MES is the gain in mutual information between the maximum $y^\star$ and the next recommendation $I(\{\mathbf{x}, y\}, y^\star | \mathcal{D})$, which can be approximated analytically under certain assumptions and after some derivations \cite{wang2017max}, being one of the advantages with respect to PES that had to use the variational inference expectation propagation algorithm, by evaluating the entropy of the GP predictive distribution:
\begin{align}
\alpha(\mathbf{x}) = I(\{\mathbf{x}, y\}, y^\star | \mathcal{D}) = H(p(y|\mathcal{D}, \mathbf{x})) - \mathbb{E}(H(p(y|\mathcal{D}, \mathbf{x}, y^\star))) \approx \\ \approx \frac{1}{K} \sum_{y^\star \in Y^\star}(\frac{\gamma_{y_*}(\mathbf{x}) \psi(\gamma_{y_*}(\mathbf{x}))}{2 \Psi(\gamma_{y_*}(\mathbf{x}))} - \log(\Psi(\gamma_{y_*}(\mathbf{x})))
)\,,
\end{align}
where $\psi(\cdot)$ is the probability density function of a normal distribution, $\Psi(\cdot)$ is the cumulative density function of a normal distribution, and $\gamma_{y^*}(\mathbf{x}) = \frac{y^* - \mu(\mathbf{x})}{\sigma(\mathbf{x})}$, with $\mu(\mathbf{x})$ and $\sigma(\mathbf{x})$ being the mean and standard deviation predicted by a Gaussian process at point $\mathbf{x}$. The expectation in is taken over $p(y^* | \mathcal{D})$, which is approximated using Monte Carlo estimation by sampling a set of $K$ function maxima and optimizing the samples. The probability in the first term, $p(y | \mathcal{D}, \mathbf{x})$, is a Gaussian distribution with mean $\mu(\mathbf{x})$ and variance $k(\mathbf{x}, \mathbf{x})$ and the probability in the second term, $p(y | \mathcal{D}, \mathbf{x}, y^*)$, is a truncated Gaussian distribution such that given $y^\star$, the distribution of $y$ must satisfy $y < y^*$ to model the dependency of the maximum value of the problem. Just this expression, obtained after assumptions and derivations \cite{wang2017max}, is the analogous in the image space $\mathcal{Y}$ from all the factors used in the case of PES and the expectation propagation algorithm that used the information about the input space $\mathcal{X}$, what makes MES easier to implement. We illustrate the truncated Gaussian distribution modeled by MES for clarity in Figure \ref{fig:trun}.

The information gained about the maximum value of the objective function is also targeted by output-space predictive entropy search, OPES \cite{hoffman2015output}. One of the advantages of this method is that it can be applied to inputs that are the union of disjoint and differently-dimensioned spaces. The theoretical acquisition function is the same one as MES, being the first term the entropy of a Gaussian random variable, the expectation approximated by Monte Carlo but in this case the entropy of the conditional predictive distribution is approximated with the expectation propagation algorithm, which makes it more difficult to implement than MES, hence having less popularity than the previous method. 

Analogously with the previous approaches, another proposal insists that sampling the global minimizer of the function $p(\mathbf{x}^\star)$ or computing the expectation propagation algorithm for each of the samples is a costly process. Consequently, the fast information theoretical based Bayesian optimization approach (FITBO) \cite{ru2018fast} tries to alleviate this issue, avoiding the need of sampling the global maximizer $p(\mathbf{x}^\star)$, also working with the output space $\mathcal{Y}$. FITBO has the same theoretical expression for the acquisition function as MES and OPES, but approximates it differently. In particular, FITBO expresses the objective function in a parabolic form $f(\mathbf{x}) = \eta + \frac{1}{2}g(\mathbf{x})^2$, where $\eta$ is a hyperparameter representing the global minimum $f^\star$. By doing this, FITBO reduces the costly process of sampling the global minimum to the cheaper process of sampling one hyperparameter, overcoming the speed bottleneck of information-theoretic approaches. The first term of the acquisition function is approximated with this approach into the entropy of a Gaussian mixture that also needs to be approximated and the second term is the expected entropy of a one-dimensional Gaussian distribution, that can be computed analytically. Details of the approximation to the theoretical acquisition function and how they dealt with the hyperparameter can be found in the paper \cite{ru2018fast}.   

Previous approaches are difficult to generalize to batch Bayesian optimization, where we want to determine which is the set of inputs $\mathbf{X} \in \mathcal{X}$ that maximize the expected information gain, mainly because of extra non-Gaussian factors in the distributions. Trusted-maximizers entropy search \cite{nguyen2021trusted} is a similar proposition as PES but it measures how much an input query contributes to the information gain on the optimum over a finite set of trusted maximizers $\mathcal{X}^\star\subseteq \mathcal{X}$ instead that in the continuous input space $\mathcal{X}$ as in PES. These trusted maximizers are inputs that are more likely to be the global maximizer $\mathbf{x}^\star$ in the current BO iteration, so if we are able to know their location, the evaluation of the rest of points would be useless, accelerating the method. $\mathcal{X}^\star$ is formed using samples of $\mathbf{x}^\star$ drawn from the posterior belief $p(\mathbf{x}^\star|\mathcal{D})$ and optimizing them. Several sampling techniques and the expectation propagation algorithm are then used to approximate the acquisition function \cite{nguyen2021trusted}.  

A rectified version of MES \cite{nguyen2022rectified} approximates the expected information gain about the maximum value with a closed-form probability density for the observation conditioned on the max-value and employ stochastic gradient ascent with reparameterization to efficiently optimize RMES. Improved and rectified versions of MES also appear in Bayesian optimization applied to complex scenarios, as we will illustrate in the following section.

Previous methods focus on gaining information about the localization of the optimum $p(\mathbf{x}^\star)$ or about the value of the maximizer $p(y)$. However, both are random variable that represent different information about the optimum and hence they can be combined in a joint probability distribution $p(\mathbf{x}^\star, y)$. This is what is done in Joint Entropy Search (JES) \cite{hvarfner2022joint}, where the recommended point $\mathbf{x}$ is going to be the one that maximizes the expected information gain about the joint probability distribution between the location of the optimum and its value given the previous observations $p(\mathbf{x}^\star, y | \mathcal{D})$. More formally, let the mutual information between the random variables $(\mathbf{x}^\star, f^\star)$ and a predicted point $(\mathbf{x}, y)$ be denoted as the joint entropy search acquisition function:
\begin{align}
 \alpha_{\text{JES}}(\mathbf{x}) &= I((\mathbf{x}, y); (\mathbf{x}^\star, f^\star) \mid \mathcal{D}) \\
    &= H[p(y \mid D, \mathbf{x})] - \mathbb{E}_{(\mathbf{x}^\star, f^\star)} \big[ H[p(y \mid D, \mathbf{x}, \mathbf{x}^\star, f^\star)] \big] \\
    &= H[p(y \mid D, \mathbf{x})] - \mathbb{E}_{(\mathbf{x}^\star, f^\star)} \big[ H[p(y \mid D \cup (\mathbf{x}^\star, f^\star), \mathbf{x}, f^\star)] \big]\,,
\end{align}
where this approach offered state-of-the-art results as the information gained about the optimum is more rich than in the case of MES and PES related methods. The $L$ samples $(\mathbf{x}^\star, f^\star)_{l=1}^L$ are obtained in a slightly modified manner than they were obtained in PES to solve the expectation with Monte Carlo. The distribution $p(y \mid D \cup (\mathbf{x}^\star, f^\star), \mathbf{x}, f^\star)$ is a truncated normal, but as the entropy is computed with respect to the density over the noisy observations $y$, they follow a extended skew distribution which does not have an analytical solution. The authors solve this issue matching the moments of that distribution with those of a truncated normal distribution over $f$, which turns to be a lower bound on the information gain, having an analytical closed-form that is used to approximate the previous acquisition function.

Further recent research considered generalizations of Shannon entropy to be used for information gain, for example from work in statistical decision theory with a class of uncertainty measures parameterized by a problem-specific loss function corresponding to a downstream task \cite{neiswanger2022generalizing}, or using $\alpha$-divergences instead of the KL divergence to derive Alpha Entropy Search (AES). 

In AES, the $\alpha$ hyperparameter of the $\alpha$ divergence is marginalized with a discrete grid of size $A$ that forms an ensemble of acquisition functions each of those parametrized with a different value of $\alpha$, being a different divergence used for the expected generalized mutual information \cite{fernandez2024alpha} of the joint probability distribution of the value $f^\star$ and the location of the optimum $\mathbf{x}^\star$ and a new evaluation $(\mathbf{x}^\star, y)$, obtaining state-of-the-art results.
\begin{align}
\alpha(\mathbf{x})_{AES} = \sum_{a=1}^A \alpha_a(\mathbf{x})\,.
\end{align}
Less formally, AES targets the same acquisition function than JES but generalizes it from the Shannon divergence into alpha divergences. The next section will introduce how we can generalize the previous approaches to more complex scenarios that the one dealt in this section. 

\subsection{Bayesian optimization applied to complex scenarios}
Until now we have described how we can use information-theoretic concepts in vanilla Bayesian optimization, where we want to obtain a configuration whose value approximates the optimum of a objective function $f: \mathcal{X} \to \mathbb{R}$. However, all the approaches presented previously can be adapted to more complex scenarios where we can also apply Bayesian optimization \cite{garrido2021advanced}. In this section, we briefly illustrate some of these approaches and the problems that they solve.

\subsubsection{Constrained Bayesian optimization}
\begin{figure}[htb!]
  \centering
  \includegraphics[width=0.99\textwidth]{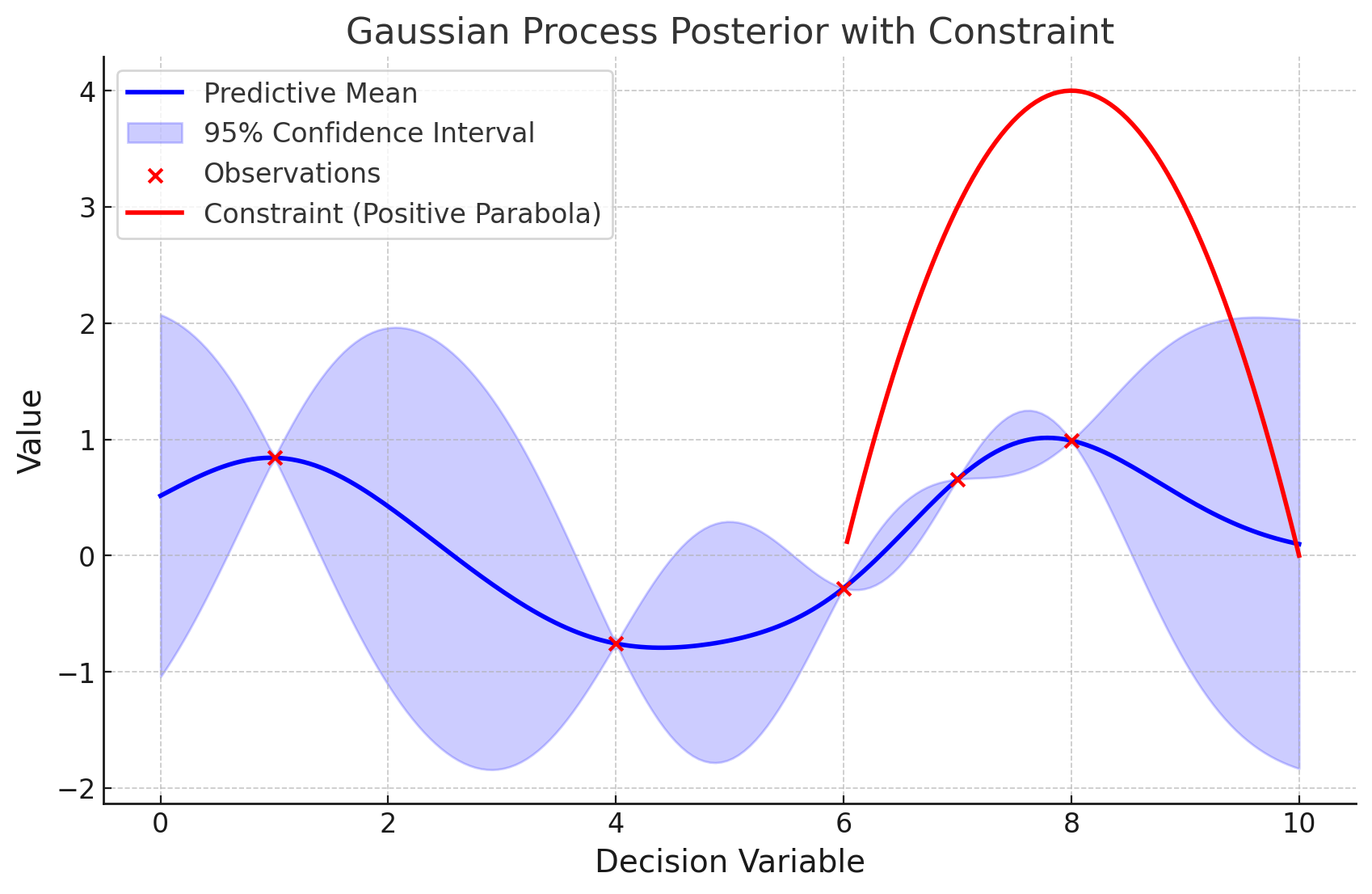}
	\caption{Constrained Black-box optimization problem. The constraint is drawn on red. In practice it is unknown and modeled with a Gaussian process. All the input space area where the constraint, in this example a parabola, is negative makes the solution infeasible, so the solution of this problem is the point that minimizes the objective such that the parabola function is positive.}
  \label{fig:constraint}
\end{figure}

Constrained Bayesian optimization only allows as solutions to the problem those points where a set of $m$ black-box constraints $\mathbf{c}(\mathbf{x})$ are validated. More formally, addresses problems of the form:
\begin{align}
\max_{\mathbf{x} \in \mathcal{X}} f(\mathbf{x}) \quad \text{subject to} \quad c_i(\mathbf{x}) \leq 0 \; \forall i \in \{1, \dots, m\},
\end{align}
where \( f(\mathbf{x}) \) is the black-box objective to be maximized, and \( c_i(\mathbf{x}) \) are black-box constraint functions usually modeled as independent from each other and the objective. We provide a visualization of an example in Figure \ref{fig:constraint}.
In this scenario, Gaussian processes, are typically used to model both \( f(\mathbf{x}) \) and \( c_i(\mathbf{x}) \), enabling probabilistic reasoning over feasible regions given the values of the predictive distributions over all the Gaussian processes conditioned by previous observations. Each black-box is typically modelled with one independent Gaussian process. The input space $\mathcal{X}$ is now constrained into the feasible space $\mathcal{F} \subset \mathcal{X}$.

We can generalize the previous approaches to this setting in different ways. For example, PES can be enhanced into the predictive entropy search with constraints (PESC) approach \cite{hernandez2015predictive}), that adds a set of delta factors $\prod_{i=1}^m \mathbb{I}[c_i(\mathbf{x}) \leq 0]$ to constrain the invalid solutions into the acquisition function, where \( \mathbb{I}[\cdot] \) is the indicator function ensuring feasibility. To solve the new acquisition function, these factors are approximated with Gaussians using the expectation propagation algorithm, adding more complexity to the approach. The MES acquisition can also be generalized into its version with constraints \cite{perrone2019constrained} approximating this time the constraints factors using the Laplace approximation. 

\subsubsection{Multi-objective scenario}
\begin{figure}[htb!]
  \centering
  \includegraphics[width=0.99\textwidth]{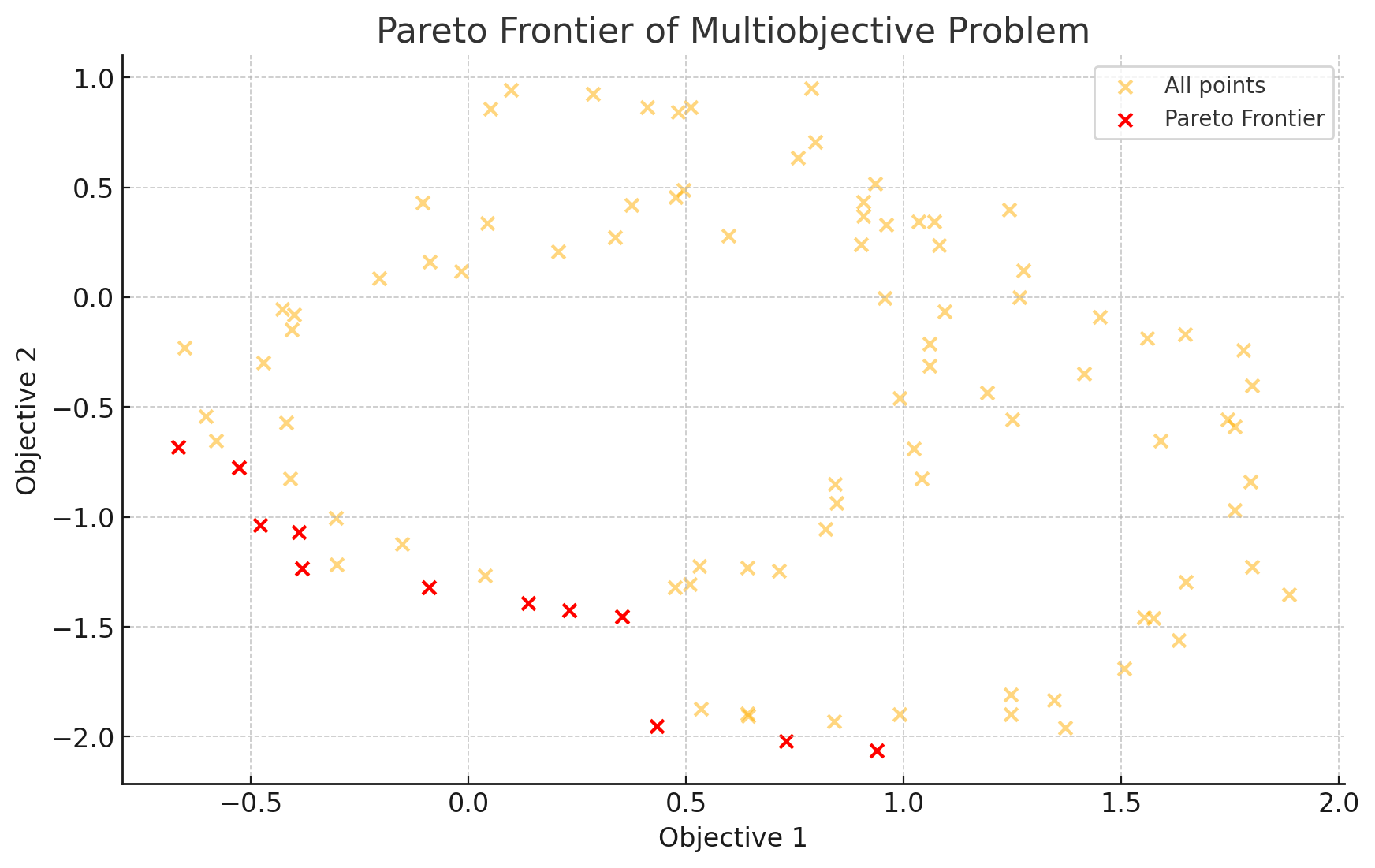}
	\caption{Visual example of the Pareto frontier, red crosses, of a two black-box optimization problem. Multi-objective Bayesian optimization like MESMO model this set as a random variable, learning information about it in an iterative fashion.}
  \label{fig:pf}
\end{figure}

Multi-objective black-box optimization focuses on solving problems where not only one but a set of conflicting black-box objectives $\mathbf{f}(\mathbf{x})$ must be satisfied. The goal is hence to approximate the Pareto frontier $\mathcal{Y}$, which represents the set of trade-off solutions where no objective can be improved without degrading another. The key concept here is that we can not model the entropy of the optimum but about different random variables associated with the solution of the multi-objective problem, such as the Pareto set $\mathcal{X}^\star$ defined as:
\begin{align}
    \mathcal{X}^\star = \left\{ \mathbf{x} \in \mathcal{X} \;\middle|\; \nexists \; \mathbf{x}' \in \mathcal{X}, \mathbf{f}(\mathbf{x}') \prec \mathbf{f}(\mathbf{x}) \right\}\,.
\end{align}
Strategies like PESMO \cite{hernandez2016predictive} and MESMO \cite{belakaria2019max} are direct generalizations of the MES and PES approaches seen in the previous section. Concretely, that guide the search by selecting evaluations that maximize the expected information gain about the Pareto set $\mathcal{X}^\star$ in the case of PES and about the Pareto frontier $\mathcal{Y}$ in the case of MES. We illustrate the Pareto frontier in Figure \ref{fig:pf}.
For example, in the case of PESMO, using the symmetric property of mutual information, we obtain that now the conditional distribution is conditioned on the Pareto set $\mathcal{X}^\star$ and the expectation is over the distribution of the Pareto set $\mathcal{X}^\star$:
\begin{align}
    \alpha(\mathbf{x}) = H(y \mid \mathcal{D}, \mathbf{x}) - \mathbb{E}_{\mathcal{X}^\star} \big[ H(y \mid \mathcal{D}, \mathbf{x}, \mathcal{X}^\star) \big]\,,
\end{align}
which adds significant complex technicalities into the final solution, as many more non-Gaussian factors need to be included in the predictive distribution to make it conditional on the Pareto set $\mathcal{X}^\star$, adding extra difficulties into the implementation that MESMO alleaviates. Analogously, we can also generalize JES \cite{tu2022joint} into this setting by acquiring information about the joint distribution of the Pareto set and frontier. The multi-objective setting has also been targeted for Bayesian optimization by other similar strategies \cite{suzuki2020multi,belakaria2021output,james2024multi}.

\subsubsection{Constrained multi-objective scenario}
We can combine the previous two scenarios into a single one, where we want to estimate the Pareto set $\mathcal{X}^\star$ of a set of conflictive black-box objectives $\mathbf{f}(\mathbf{x})$ such that they satisfy a set of black-box constraints $\mathbf{c}(\mathbf{x})$. That is:
\begin{align}
    \min_{\mathbf{x} \in X} & \quad f_1(\mathbf{x}), \dots, f_K(\mathbf{x}) \\
    \text{s.t.} & \quad c_1(\mathbf{x}) \geq 0, \dots, c_C(\mathbf{x}) \geq 0\,,
\end{align}
where $k$ is the number of black-box objectives and $c$ is the number of black-box constraints. Hence, the would like to estimate the feasible Pareto set $\mathcal{F}^\star$. Each black-box can be modelled using a Gaussian process that is going to be conditioned with the observations of the problem $\mathcal{D}$. Previous approaches can be generalized towards this problem such as a generalization of PES called predictive entropy search with multiple objectives and constraints, PESMOC \cite{garrido2019predictive} that search for the location of the feasible Pareto set by adding together the non-Gaussian factors belonging to the validation of the constraints and the validation of a point belonging to the Pareto set that are approximated as Gaussians using the expectation propagation algorithm, which makes the approach engineeringly very difficult to implement. Analogously, we find in the literature a generalization of MES, called max-value entropy search for multiple objectives and constraints, MESMOC \cite{belakaria2020max}, that is easier to implement than PESMOC as the approximation is done analytically. In particular, in MESMOC+ \cite{fernandez2023improved} the improved version of MESMOC, approximates the following expression with procedures that are explained in the paper:
\begin{align*}
\alpha(\mathbf{x}) &= H\bigl(\mathbf{y} \mid \mathcal{D}, \mathbf{x}\bigr) 
            - \mathbb{E}_{\mathcal{Y}\star}\Bigl[ H\bigl(\mathbf{y} \mid \mathcal{D}, \mathbf{x}, \mathcal{Y}^\star\bigr) \Bigr]\,,
\end{align*}
where $\mathcal{Y}\star$ is the feasible Pareto frontier of the constrained multi-objective problem, $\mathbf{x}$ is an input space point, $\mathbf{y}$ is the evaluation of the input space point in the image space and $\mathcal{D}$ is the dataset of previous observations. We strongly recommend the reader to study carefully how this acquisition function is approximated in the references, as it is not an easy process, involving several steps.

\subsubsection{Parallel Bayesian optimization}
\begin{figure}[htb!]
  \centering
  \includegraphics[width=0.99\textwidth]{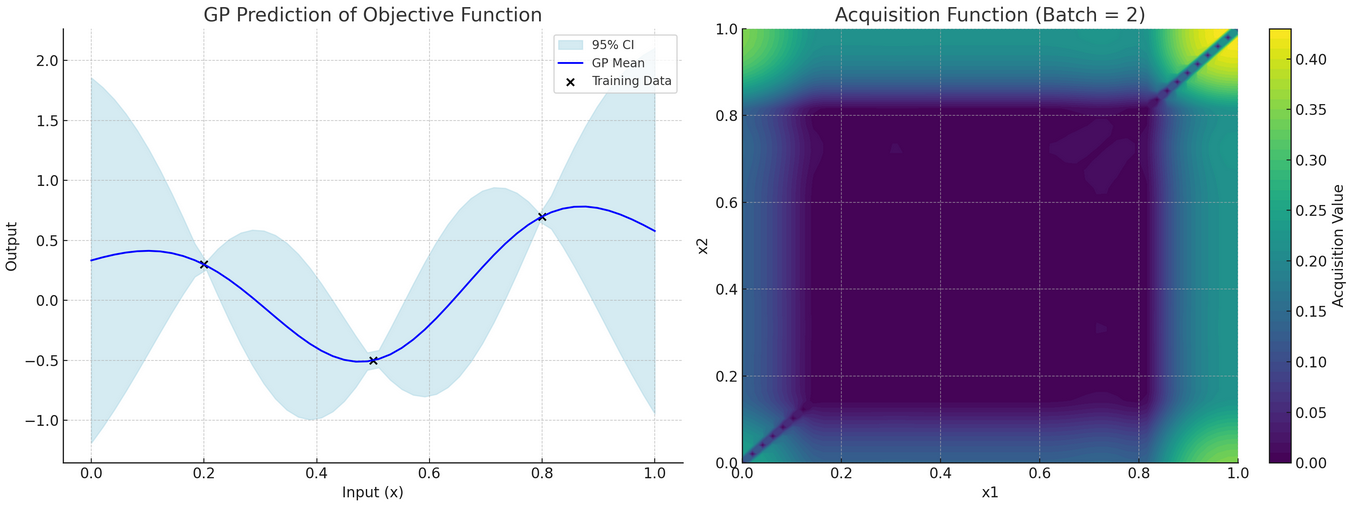}
	\caption{Batch Bayesian optimization of a one-dimensional objective function with batch size equals to 2. The plot of the acquisition is illustrating the expected utility of evaluating every possible pair of points of the input space. The values of the diagonal are low because evaluating the same point twice does not give you any additional information with respect to evaluating the point a single time in a noiseless setting.}
  \label{fig:bbo}
\end{figure}

In previous subsections we are always interested in selecting the point $\mathbf{x}$ that maximizes an information-theoretic metric towards some random variable that is related to the optimizer of the black-box function $f(\mathbf{x})$. This relies on the assumption that we only have one sequential evaluator of the objective function. However, this does not need to be the case, as for example in hyper-parameter tuning we may have a computing cluster. Hence, it is also interesting to suggest a batch of $L$ points $\mathbf{X}$ that maximizes the information gain about the optimum $\mathbf{x}^\star$. We illustrate this idea in Figure \ref{fig:bbo}.
Several approaches have extended the previous approaches into the parallel setting as parallel predictive entropy search, PPES \cite{shah2015parallel}, the batch version of PES. Also, this approach can be extended to solve the constrained multi-objective setting introduced previously, as parallel predictive entropy search for multi-objective Bayesian optimization with constraints, PPESMOC \cite{garrido2023parallel}, targets, significantly incrementing the difficulty of implementing it but successfully solving the problem. In particular, PPESMOC solves this problem by approximating the following acquisition function:
\begin{align}
    \alpha(\mathbf{X}) &= H[p(\mathbf{Y} \mid \mathcal{D}, \mathbf{X})] 
    - \mathbb{E}_{p(\mathcal{X}^\ast \mid \mathcal{D})} 
    \big[ H[p(\mathbf{Y} \mid \mathcal{D}, \mathbf{X}, \mathcal{X}^\ast)] \big]\,,
\end{align}
where $\mathbf{X}$ is a set of $B$ input space points such that $B$ is the number of points of the batch, $\mathcal{X}^\ast$ is the feasible Pareto set, $\mathcal{D}$ is the dataset of previous evaluations and $\mathbf{Y}$ is the evaluation of $\mathbf{X}$ under the black boxes and batches of points. As it can be seen, as the problem is more complex it involve more complexity of the acquisition functions, full of factors to approximate, which makes these approaches powerful but very difficult to implement.

\subsection{Other complex scenarios}
In this subsection, we discuss more advanced Bayesian optimization methods that have not been dealt before by focusing on specialized scenarios such as partial evaluations, input noise, multi-fidelity expansions, and multi-objective or high-dimensional tasks that have been tackled using an information-theoretic approach, as in the previous subsections. 

Freeze-thaw Bayesian optimization \cite{swersky2014freeze} tackles the challenge of incomplete or interrupted evaluations by dynamically finishing the experiments if they are not going to be worth as a result of loss machine learning curves. It can be done by assessing the information of the curve, whether it is expected to provide better performance or not. Noisy input entropy search \cite{frohlich2020noisy} extends classical entropy-based criteria by including noisy input variables, refining the entropy search acquisition function with an explicit modeling of input uncertainty to maintain robust predictions. Multi-fidelity MES parallel \cite{takeno2020multi} is an extension of MES to address multi-fidelity scenarios, where some configurations are evaluated using less resources and if they are successful then they are evaluated using more to save resources, that is the idea of multiple fidelities. Alternatively, multi-fidelity multi-objective OES \cite{belakaria2020multi} further generalizes this framework by incorporating multiple objectives, also extending the entropy search approach. All the approaches seen until now work successfully for less than 8 dimensions, having trouble with a higher number of dimensions, consequently, high-dimensional entropy search \cite{li2021active} tackles the curse of dimensionality by employing scalable approximations that retain an information-theoretic perspective on exploration and exploitation in high dimensional input spaces. Multi-agent Bayesian optimization \cite{ma2023gaussian} introduces decentralized decision-making among coordinated agents, enabling simultaneous exploration with localized updates and a global consensus mechanism that generates collective learning. Multi-objective scenarios seen until now assume independence between the conflictive objectives, but usually they are negatively correlated, henc,e multi-task entropy search \cite{moss2021mumbo} exploits correlations across related tasks to share knowledge and speed up convergence, using task-specific covariance structures to guide sampling policies. High-dimensional many-objective entropy search \cite{bian2023bayesian} enhances the high-dimensional approach to tackle not only two or three but more objectives in a high-dimensional input space. 

\section{Conclusions and Further Lines of Research}
This document has shown a comprehensive tutorial and survey on how modern information-theoretic Bayesian optimization approaches work, covering their methodological evolution in a historical fashion after illustrating the basic information-theoretic concepts. We have also illustrated the versatility of these methodologies across a diverse range of complex Bayesian optimization scenarios, like the constrained multi-objective scenario, emphasizing the adaptability of information-theoretical Bayesian optimization to complex problem settings. As these approaches are difficult to implement in practice, to support further exploration, we have provided the references for each approach, where all the technicalities are explained properly.

There are several future research directions that are worth exploring for advancing this field. Key areas include developing improved approximations for the entropy search acquisition functions, such as using approaches as Power Expectation Propagation \cite{minka2004power}, or exploring generalized notions of entropy, like the Sharma-Mittal entropy, to develop more general acquisition functions \cite{amigo2018brief}. In order to determine the hyper-parameters that those acquisition functions are going to have, meta Bayesian optimization approaches, including bandits and Bayesian model selection, represent another line for innovation in designing information-theoretic acquisition functions. Additionally, as they are going to appear a lot of acquisition functions based on those concepts, it is going to be useful to study theoretical guarantees of convergence of these acquisitions, to determine for example upper bounds on the cumulative regret of these approaches. Another interesting research direction includes adapting and selecting probabilistic surrogate models for specific problems, including ensembles of them. Finally, enhancing scalability through information compression techniques of the evaluations to compete with metaheuristics in problems where cheaper evaluations can be done is also a setting worth to explore.
\bibliographystyle{abbrv}
\bibliography{references}

\end{document}